\documentclass{article}

\usepackage{microtype}
\usepackage{graphicx}
\usepackage{subcaption}
\usepackage{booktabs} 

\usepackage{algorithm}
\usepackage{enumitem}
\usepackage{hyperref}
\usepackage[T1]{fontenc}

\usepackage[preprint]{icml2026}

\let\algorithmicrequire\relax
\let\algorithmicensure\relax
\let\algorithmiccomment\relax
\let\algorithmicend\relax
\let\algorithmicif\relax
\let\algorithmicthen\relax
\let\algorithmicelse\relax
\let\algorithmicfor\relax
\let\algorithmicforall\relax
\let\algorithmicdo\relax
\let\algorithmicwhile\relax
\let\algorithmicloop\relax
\let\algorithmicrepeat\relax
\let\algorithmicuntil\relax
\let\algorithmicfunction\relax

\usepackage{algpseudocode}

\usepackage{amsmath}
\usepackage{amssymb}
\usepackage{mathtools}
\usepackage{amsthm}

\usepackage[capitalize,noabbrev]{cleveref}

\theoremstyle{plain}
\newtheorem{theorem}{Theorem}[section]
\newtheorem{proposition}[theorem]{Proposition}

\theoremstyle{definition}
\newtheorem{definition}[theorem]{Definition}

\theoremstyle{remark}
\newtheorem{remark}[theorem]{Remark}

\usepackage[textsize=tiny]{todonotes}

\newcommand{\E}{\mathbb{E}}
\newcommand{\Prob}{\mathbb{P}}

\crefname{section}{Sec.}{Secs.}
\Crefname{section}{Section}{Sections}

\crefname{appendix}{App.}{Apps.}
\Crefname{appendix}{Appendix}{Appendixes}

\icmltitlerunning{Semantic Collisions and the Scale-Dependence of Effective Duplicates}

\begin{document}

\twocolumn[
\icmltitle{Scale Dependent Data Duplication}

\icmlsetsymbol{equal}{*}

\begin{icmlauthorlist}
\icmlauthor{Joshua Kazdan$^*$}{stanstat}
\icmlauthor{Noam Levi$^*$}{epflai}
\icmlauthor{Rylan Schaeffer}{stan}
\icmlauthor{Jessica Chudnovsky}{stan}
\icmlauthor{Abhay Puri}{sn}
\icmlauthor{Bo He}{imc}
\icmlauthor{Mehmet Donmez}{imc}
\icmlauthor{Sanmi Koyejo$^\dag$}{stan}
\icmlauthor{David Donoho$^\dag$}{stanstat}
\end{icmlauthorlist}

\icmlaffiliation{stanstat}{Department of Statistics, Stanford University}
\icmlaffiliation{epflai}{AI4Science, EPFL}
\icmlaffiliation{stan}{Department of Computer Science, Stanford University}
\icmlaffiliation{sn}{ServiceNow Research}
\icmlaffiliation{imc}{IMC Trading}

\icmlcorrespondingauthor{Joshua Kazdan}{jkazdan@stanford.edu}

\vskip 0.1in
]
\renewcommand{\thefootnote}{\fnsymbol{footnote}}
\footnotetext[1]{Equal contribution}
\footnotetext[2]{Equal advising}
\renewcommand{\thefootnote}{\arabic{footnote}}

\printAffiliationsAndNotice{}  

\begin{abstract}

Data duplication during pretraining can degrade generalization and lead to memorization, motivating aggressive deduplication pipelines. However, at web scale, it is unclear what constitutes a ``duplicate'': beyond surface-form matches, semantically equivalent  documents (e.g.\ translations) may induce redundant training signals once models become sufficiently capable.  Practically, this means that semantic duplicates operate increasingly like exact duplicates during training. We present evidence that duplication is scale-dependent in two ways. First, as model capability increases, cross-entropy loss gradients for semantically equivalent documents become more aligned. Smaller models, by contrast, produce gradients that reflect surface similarity (e.g., shared tokens) rather than semantic similarity. 
Second, we embedded all 192 million FineWeb-Edu-Dedup documents using EmbeddingGemma-300m. For moderate corpus sizes, the cosine similarity between nearest-neighbors follows an isotropic power law baseline. However, as corpus size grows to hundreds of billions of tokens, the nearest-neighbor similarities deviate sharply, indicating accelerated semantic collisions.
Finally, controlled pretraining on data sampled with replacement from pools of finite unique documents shows that limited uniqueness yields mild degradation for small models, but rapidly increasing loss penalties for larger models, breaking naive scaling extrapolation.  We derive explicit scaling laws that allow practitioners to estimate deviation from expected scaling due to limited semantic uniqueness of the pretraining corpus. Our results identify and resolve an unstudied source of scale-dependence, allowing for more accurate prediction at scale.


\end{abstract}

\section{Introduction}

Modern language models scale by increasing parameters, compute, and training tokens.
For example, Llama 1 \citep{touvron2023llamaopenefficientfoundation} trained on $\sim$1T tokens, while the Llama4 herd \citep{adcock2026llama4herdarchitecture} trained on up to 40T tokens.  
At these scales, even small fractions of duplicated data can materially reduce the number of distinct training examples and harm downstream performance, emphasizing the importance of deduplication \citep{carlini2021extractingtrainingdatalarge, hernandez2022scalinglawsinterpretabilitylearning, lee2022deduplicatingtrainingdatamakes, comanici2025gemini25pushingfrontier}.

Deduplication is often framed as a dataset property: to deduplicate, simply remove exact duplicates and near-duplicates using simhashing techniques \cite{simhash1997, manku_minhash, khan2025lshbloommemoryefficientextremescaledocument, lee2022deduplicatingtrainingdatamakes}. Yet, what practically counts as a ``duplicate'' depends on the model as well: two documents that appear distinct may, to a sufficiently capable model, provide redundant training signal, and thus degrade training just as exact duplicates would. 
This work identifies a previously unknown source of scale dependence: as models become more capable, \emph{semantic duplicates} induce the same gradients during training. In tandem, capable models are trained on larger corpora, in which the number of semantic collisions rapidly increases. Together, these effects create a recipe for model degradation.




\textbf{Contributions: }
\begin{enumerate}[itemsep=1pt]
    \item We quantify the \textbf{emergence of semantic sensitivity} during training by measuring cosine similarity between per-document cross-entropy gradients across a suite of models and semantic-preserving transformations.  We find that in more capable models, semantic duplicates induce similar gradients during training.
    \item We study \textbf{semantic collisions} by embedding 192M documents from FineWeb-Edu-Dedup \citep{penedo2024finewebdatasetsdecantingweb} documents and analyzing nearest-neighbor (NN) statistics across dataset scales from $10^4$ to $10^8$ documents.  We discover that power laws governing scaling for moderate corpus sizes break down for large corpora.  This collapse of scaling laws occurs earlier for synthetic corpora, revealing lower semantic diversity.
    \item 
    We examine the consequences for predictability by training scaling ladders on streams sampled with replacement from finite pools of $K$ unique documents, showing that limited uniqueness breaks naive scaling extrapolation. We derive more complete scaling laws that explicitly quantify the effects of limited uniqueness, restoring predictability. Furthermore, we show how to estimate an effective $K$ directly from mean nearest-neighbor cosine similarity.
\end{enumerate}

We defer related work to Appendix \ref{related_work}.

\begin{figure*}[t]
    \centering
    \includegraphics[width=0.95\linewidth]{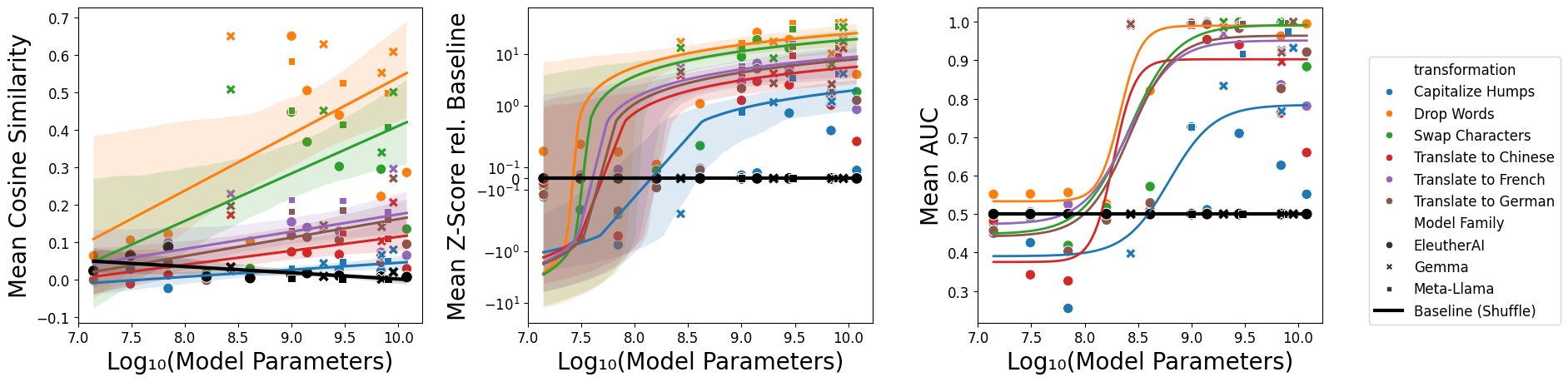}
    \caption{
    \textbf{Semantic-preserving transformations yield more aligned gradients for larger/stronger models.}
    We sample $N{=}1000$ FineWeb-Edu-Dedup documents and compute per-document gradients of normalized next-token cross-entropy (Eq.~\ref{eq:doc-loss}) for each model.
    We report mean cosine similarity between (i) unrelated document pairs (negative baseline) and
    (ii) each document and its transformed counterpart (positives), including translations and light surface perturbations.
    Smaller/weaker models exhibit gradient similarity dominated by surface cues (language/casing), often failing to separate positives from negatives.
    As capability increases, positives become consistently more aligned than the negative baseline.
    Error bars show per-document standard deviation.  Per-model-family results are in \Cref{fig:grad-sim-ind-family}.
    }
    \label{fig:grad-sim-matrix}
\end{figure*}

\section{Emergence of Semantics}
\label{sec:emergence}

As model capabilities increase, semantically equivalent documents induce similar training signals, as measured by the gradient of the per-document cross-entropy loss.
Consequently, if two documents are semantic duplicates (e.g., translations), then a sufficiently capable model will update its parameters in similar directions when trained on both documents.  Practically, this means that semantic duplicates operate increasingly like exact duplicates during training.  

\subsection{Experimental Setup}
\label{subsec:grad-setup}

We sample $N=1000$ texts $\{x_i\}_{i=1}^N$ from FineWeb-Edu-Dedup \citep{penedo2024finewebdatasetsdecantingweb}.
To reduce variance due to length, each text is truncated to at most $T=2000$ tokens using the tokenizer of the model under evaluation.

We compute the per-document full-parameter gradient
\begin{align}
\label{eq:theory_doc_loss}
g(x;\theta) &= \nabla_\theta \ell(x_i; \theta),
\end{align}
where $\ell$ is the mean next-token cross-entropy:
\begin{equation}
\label{eq:doc-loss}
\ell(x;\theta)
=
\frac{1}{|x|}\sum_{u=1}^{|x|}
\mathrm{CE}\!\left(f_{\theta}(x)_{u}, x_{u+1}\right).
\end{equation}
To establish a null baseline, we sample unrelated English documents $(x_i,x_j), \quad i\neq j$ and compute cosine similarity
\begin{equation}
\label{eq:cosine}
\mathrm{sim}(x_i, x_j)
=
\frac{\langle g(x_i;\theta), g(x_j;\theta)\rangle}{\|g(x_i;\theta)\|_2\,\|g(x_j;\theta)\|_2}.
\end{equation}
We repeat this across many random pairings to estimate the baseline mean $\mu^{-}$ and standard deviation $\sigma^{-}$.

\textbf{Transformations.}
We construct a set of transformations $\mathcal{T}=\{\tau_1,\dots,\tau_L\}$ intended to preserve semantic content while perturbing surface form:

\begin{itemize}[itemsep=0.5pt]
    \item Swap Characters: with probability $0.05$, randomly replace each ascii character with another.
    \item Drop Words: Randomly delete each word with probability $0.05$.
    \item Capitalize Humps: Capitalize every other character.
    \item Translate to Chinese/French/German.
\end{itemize}
For translations, we use Google's Translate API \citep{google_translate_api}.

For each document $x_i$ and transformation $\tau$, we compute
\begin{equation}
s_i^{+}(\tau) \coloneqq \mathrm{sim}_\theta\!\left(x_i,\; \tau(x_i)\right).
\end{equation}

\textbf{Separability Metrics (Z-scores and AUC).}
To summarize separation between positives and negatives, we define:
\begin{align}
\label{eq:zscore}
z(\tau)
&\coloneqq
\frac{\mu^{+}(\tau)-\mu^{-}}{\sigma^{-}},\\
\mu^{+}(\tau)
&\coloneqq
\frac{1}{N}\sum_{i=1}^{N}
s_i^+(\tau),\\
\mu^{-}
&\coloneqq
\frac{1}{|\mathcal{S}^{-}|}
\sum_{(i,j)\in \mathcal{S}^{-}}
\mathrm{sim}_\theta\!\left(x_i,\; x_j\right),\\
(\sigma^{-})^{2}
&\coloneqq
\mathrm{Var}_{(i,j)\in \mathcal{S}^{-}}
\left[
\mathrm{sim}_\theta\!\left(x_i,\; x_j\right)
\right].
\end{align}
We also report AUC for distinguishing transformed gradients $\{g_\theta(\tau(x_i))\}$ (positives) from unrelated gradients $\{g(x_j;\theta)\}$ (negatives) using the score $\mathrm{sim}(x_i,\cdot)$.

\begin{figure}[t]
    \centering
    \includegraphics[width=\linewidth]{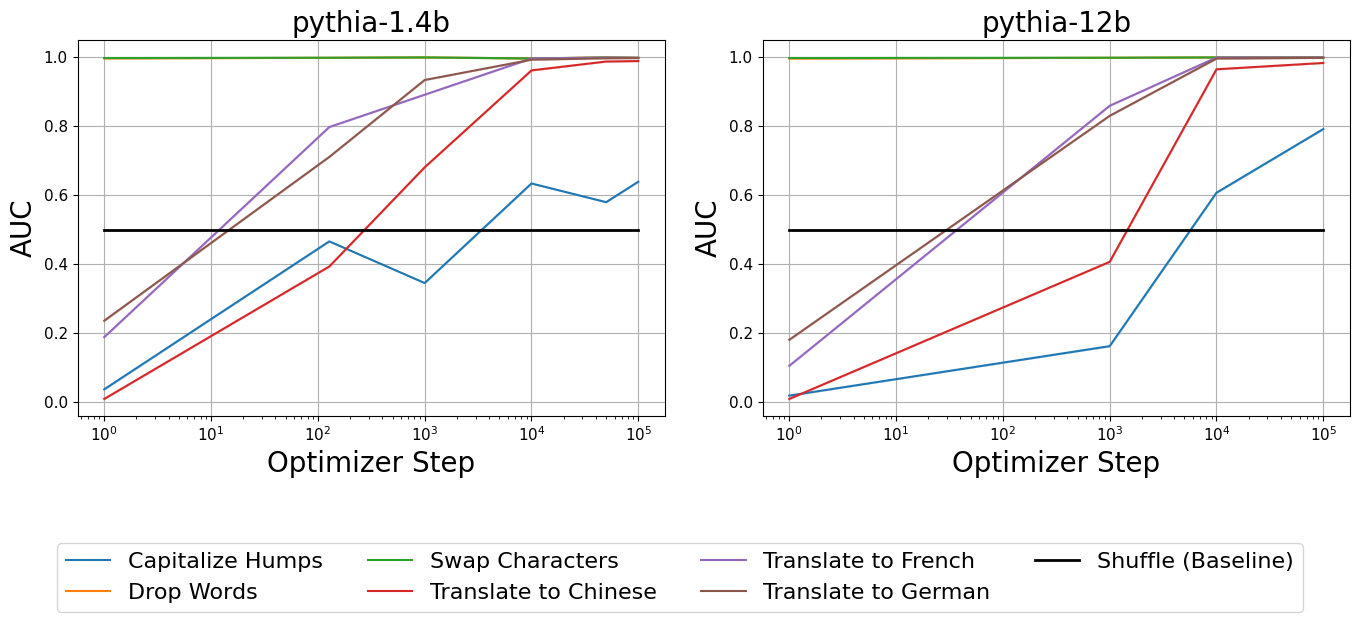}
    \caption{
    \textbf{Semantic sensitivity emerges over training and is accelerated by scale.}
    For a fixed model family, we compute AUC to detect whether a candidate gradient corresponds to a semantic-preserving transformation of the same document versus an unrelated document, with cosine similarity to the original document gradient as the score.
    Early in training, AUC remains near $0.5$ because gradients are dominated by surface-form features (language/casing).
    With additional optimizer steps, AUC increases, indicating that gradients increasingly reflect semantic content.
    Larger models reach a given AUC with fewer steps.}
    \label{fig:grad-auc-vs-steps}
\end{figure}

\subsection{Results}
\label{subsec:grad-results}

\Cref{fig:grad-sim-matrix} reports mean gradient cosine similarities for both unrelated document pairs (negative baseline) and semantic-preserving transformations (positives).
For smaller/weaker models, positive similarities for several transformations are comparable to or below the negative baseline, indicating that gradient direction is dominated by superficial features (e.g., language identity or capitalization).
As model capability increases, transformed counterparts become consistently more aligned than unrelated pairs.

To quantify separability, we compute $z(\tau)$ in Eq.~\eqref{eq:zscore} and AUC for the binary task described above.
\Cref{fig:grad-auc-vs-steps} further shows that AUC increases with training progress for a fixed family and is achieved earlier by larger models.

\textbf{Interpretation.}
Our findings suggest that semantic and exact duplicates have similar training impacts on capable models: if a model encodes meaning robustly, two semantically equivalent documents generate aligned weight updates.
This provides a mechanism by which the \emph{same dataset} can have a smaller effective size for more capable models.

\section{Semantic Collisions}
\label{sec:collisions}

When training models compute-optimally, corpus size grows in tandem with the number of parameters and  model capabilities.  In this section, we quantify the number of semantic collisions that occur in a deduplicated corpus of a given magnitude.  We find that the rate of near-duplicates follows a predictable scaling law before increasing exponentially.  

\textbf{Collision metrics.}
For a set of unit-normalized embeddings $\{v_i\}_{i=1}^{N}$, define nearest-neighbor (NN) similarity
\[
M_i \coloneqq \max_{j\neq i}\langle v_i, v_j\rangle
\quad\text{and cosine gap}\quad
\Delta_i \coloneqq 1 - M_i.
\]
We report (i) estimates of $\E[M_i]$ as a function of $N$, and (ii) tail probabilities $\Prob(M_i \ge T)$ for fixed thresholds $T$.

\subsection{Experimental Setup}
\label{subsec:nn-setup}

We embed 190M texts from FineWeb-Edu-Dedup \citep{penedo2024finewebdatasetsdecantingweb} using EmbeddingGemma-300m \citep{vera2025embeddinggemmapowerfullightweighttext}.
EmbeddingGemma-300m is a Matryoshka Representation Learning \citep{kusupati2024matryoshkarepresentationlearning} model that produces embeddings of four nested sizes (768, 512, 256, and 128); sub-embeddings are obtained by slicing and re-normalizing.

We sample subsets of embeddings with cardinality ranging from $10^4$ to $10^8$ and estimate NN cosine similarities within each pool using FAISS \citep{douze2024faiss}.

\begin{figure*}[t]
    \centering
    \includegraphics[width=0.75\linewidth]{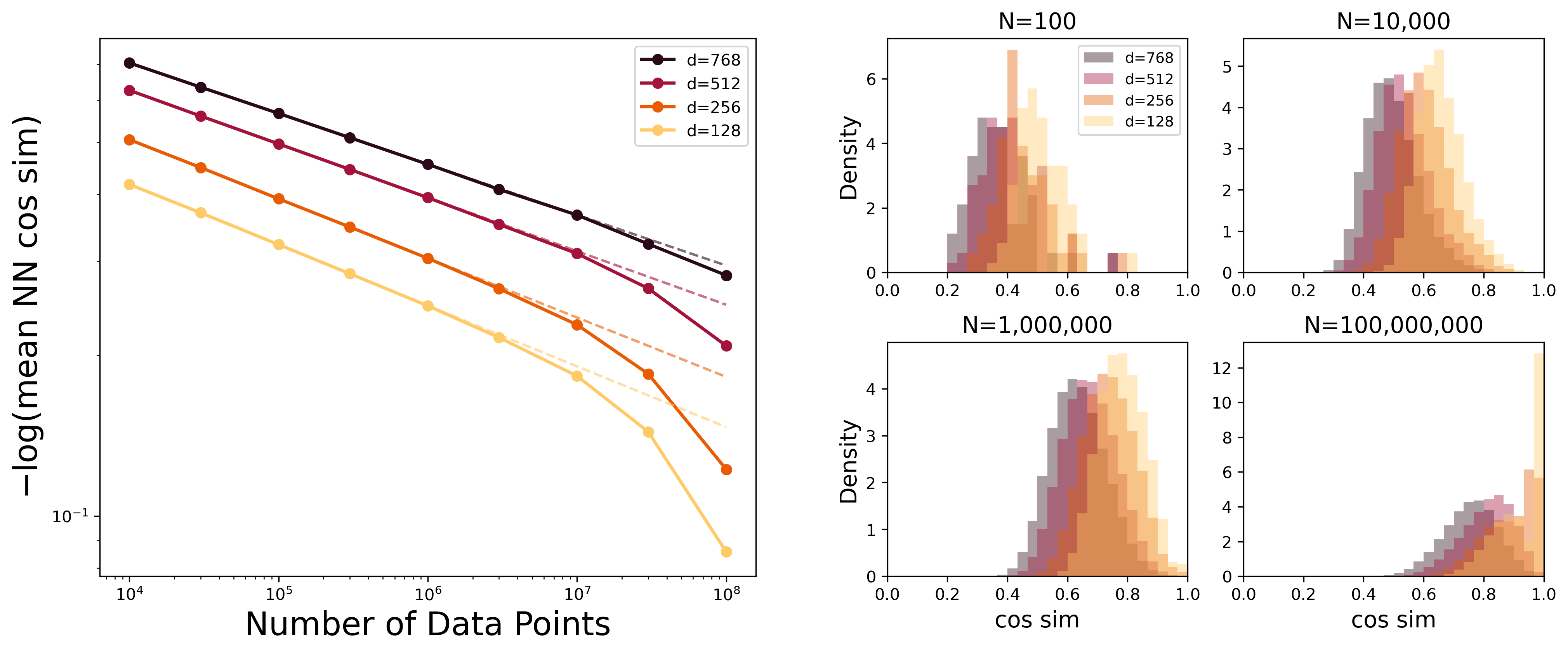}
    \caption{
    \textbf{NN cosine similarity scaling deviates sharply at large corpus sizes.}
    We embed 190M FineWeb-Edu-Dedup documents with EmbeddingGemma-300m and sample subsets of size ranging from $10^4$-$10^8$ without replacement.
    For each $N$, we estimate the mean nearest-neighbor cosine similarity using FAISS.
    Dashed lines show best-fit power laws over the small-$N$ regime where the uniform/vMF null predicts $\E[\Delta_i]\propto N^{-2/d}$.
    Beyond a scale threshold, the empirical curve steepens (smaller gaps than predicted), indicating substantially more near neighbors than expected under isotropic baselines.
    }
    \label{fig:nn-gap-scaling}
\end{figure*}

\begin{figure}
    \centering
    \includegraphics[width=\linewidth]{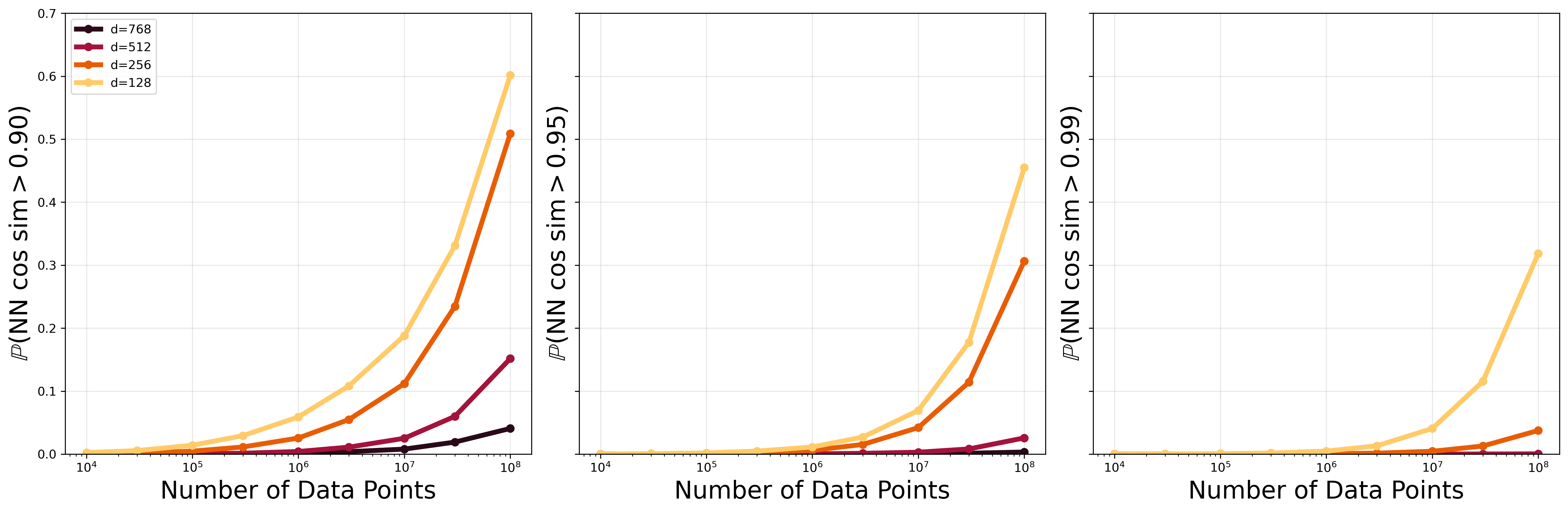}
    \caption{
    \textbf{Tail collision rates accelerate with dataset size.}
    For fixed thresholds $T$, we estimate the fraction of points with nearest-neighbor similarity $M_i \ge T$.  These increase exponentially, as predicted under an isotropic baseline.
    }
    \label{fig:nn-threshold-frac}
    \vspace{-10pt}
\end{figure}

\subsection{Results}
\label{subsec:nn-results}

\Cref{fig:nn-gap-scaling,fig:nn-threshold-frac} show that nearest-neighbor collision statistics initially match a power law, but deviate sharply at larger dataset sizes.  Collisions occur more quickly in smaller embedding spaces, as expected.  
Beyond a scale threshold, the mean cosine gap decreases \emph{faster than any fitted power law calibrated on smaller $N$}.
In log-linear coordinates, the decrease in NN cosine similarity is approximately linear over document corpus sizes less than 1M, before decaying much more quickly for corpora with over 10M documents.  This presents a potentially compound threat to language models trained at scale: larger models that are more capable of identifying semantic duplicates are trained on more data, which contains more semantic duplicates than log-linear scaling laws would predict.  Thus, models for which semantic duplicates are recognizable also experience far more of these duplicates, which could lead to loss of predictable scaling.  

\begin{figure}[t]
    \centering
    \includegraphics[width=.9\linewidth]{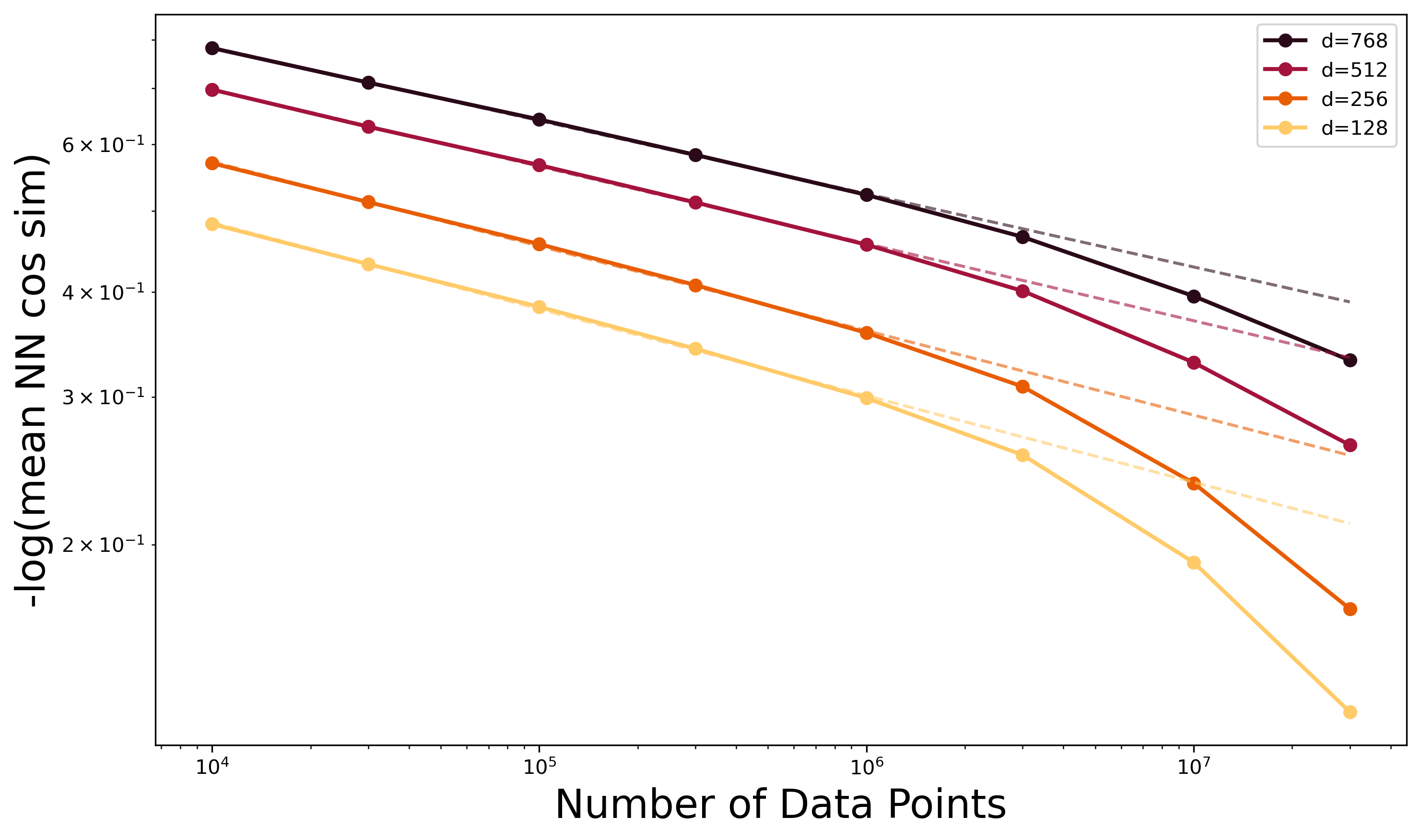}
    \caption{\textbf{Nearest-neighbor cosine similarity scaling laws collapse an order of magnitude earlier for synthetic datasets:}  We embed the fully-synthetic pretraining dataset Recycling-the-Web \citep{nguyen2025recyclingwebmethodenhance} and find that the scaling law discovered in \Cref{fig:nn-gap-scaling} occurs an order of magnitude earlier for synthetic data, suggesting that the diversity of synthetic pretraining datasets should be improved. }
    \label{fig:synthetic_diversity}
\end{figure}

\textbf{A Note on Synthetic Data: }  Recently, synthetic data has become a popular supplement for real data during pretraining and continued pretraining \citep{Mishra_2022_CVPR, chen2024diversitysyntheticdataimpact, yang2024syntheticcontinuedpretraining, kang2025demystifyingsyntheticdatallm, qin2025scalinglawssyntheticdata}, though questions remain about whether it has sufficient diversity to provide a future alternative for real data.  We repeat the experiment described in Section~\ref{subsec:nn-setup} for the fully-synthetic, 44M-document Recycling-the-Web pretraining corpus \citep{nguyen2025recyclingwebmethodenhance}.  We find that divergence from power law scaling (\Cref{fig:nn-gap-scaling}) appears an order of magnitude earlier for synthetic pretraining data (\Cref{fig:synthetic_diversity}).  

\section{Impact on Training}
\label{sec:impact}

We now probe practical implications.  If semantic collisions reduce effective uniqueness, scaling-ladder extrapolation can fail.
Because we cannot train models at the scale where \emph{semantic} duplicates are recognized in our controlled setting, we model semantic collisions via \emph{exact} document repeats (sampling with replacement), which provides a pessimistic, worst-case proxy for repeated training signals.

\subsection{Experimental Setup}
\label{subsec:impact-setup}

We sample pools of unique data of size $K$ ranging from $10^5$ through $10^8$ unique documents sampled from FineWeb-Edu-Dedup.
We construct training streams by sampling with replacement from each pool, inducing exact repeats.
As a reference, we also train on streams constructed to minimize repeats (``approximately infinite unique data'') by sampling without replacement from FineWeb-Edu-Dedup.

We train scaling ladders of decoder-only, Chinchilla-optimal \citep{hoffmann2022trainingcomputeoptimallargelanguage} transformers based on the Qwen architecture ranging from 34M--344M parameters \citep{qwen2025qwen25technicalreport, yang2025qwen3technicalreport}.
We match runs by compute (FLOPs) and report train and validation cross-entropy.


\begin{figure*}[t!]
    \centering
    \includegraphics[width=0.75\linewidth]{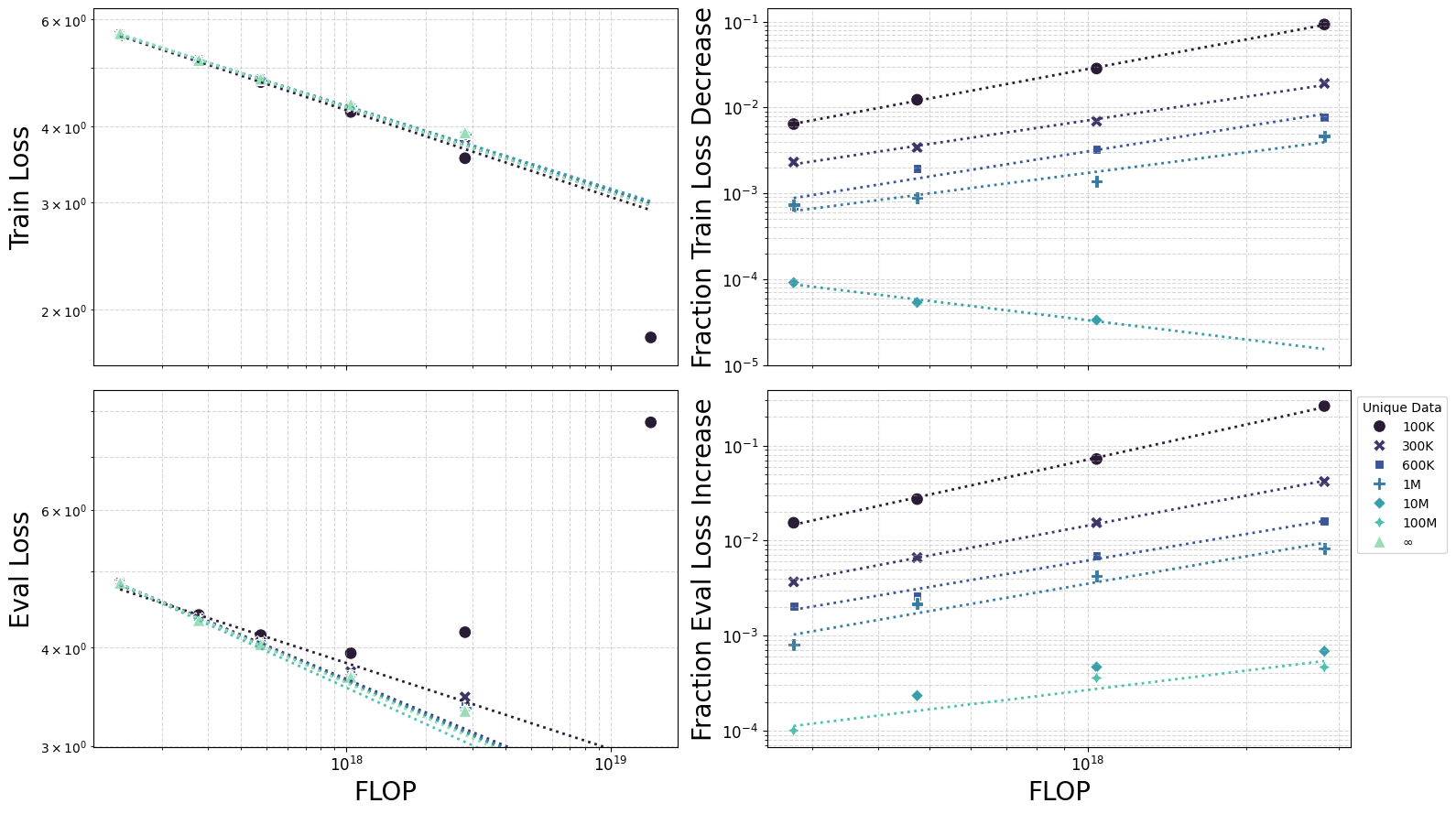}
    \vspace{-8pt}  %
    \includegraphics[width=0.83\linewidth]{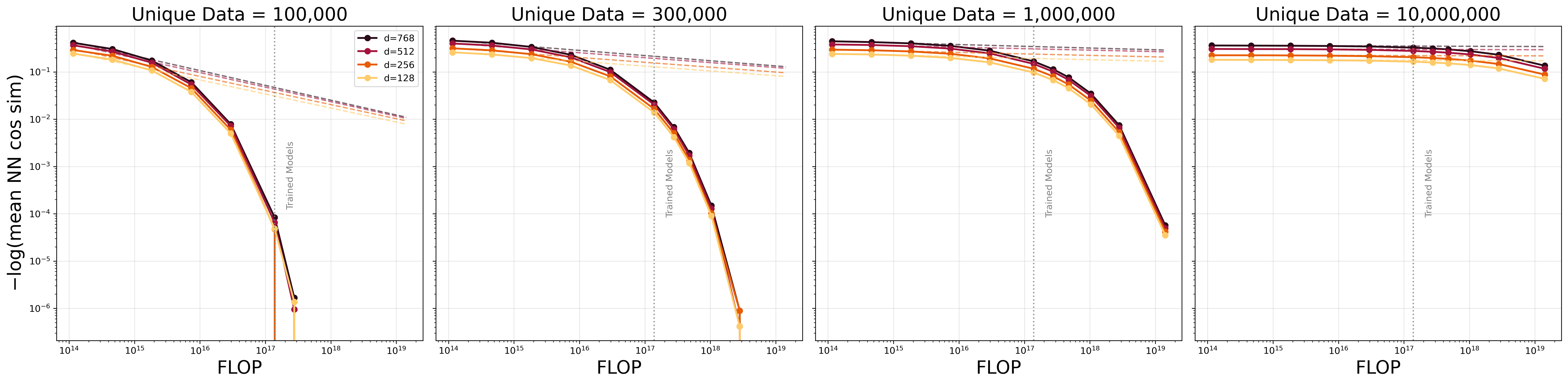}
    \vspace{2pt}  %
    \caption{
    \textbf{Finite unique data pools induce scale-dependent degradation and break naive scaling extrapolation.}
    We train model ladders at matched compute while sampling training documents with replacement from pools of size $K$ (exact repeats allowed).
    We compare against an approximately-infinite baseline with negligible repeats.
    Left: train and validation loss versus compute/scale for each $K$.
    Right: fractional loss change relative to the baseline (Eq.~\ref{eq:frac-inc}).
    Small models scale normally under small $K$, while larger models exhibit rapidly increasing penalties, implying that scaling ladders can underestimate main-run loss when effective uniqueness is limited.
    }
    \label{fig:train_eval_predictable_scaling_breakdown}
\end{figure*}


\subsection{Results and Discussion}
\label{subsec:impact-results}

\Cref{fig:train_eval_predictable_scaling_breakdown} shows that limiting $K$ produces a scale-dependent degradation pattern.
For smaller models, train and validation losses are consistent with standard scaling extrapolations, even when $K$ is small; this can mislead scaling-ladder planning.
For larger FLOP budgets, finite-$K$ streams yield increasing loss penalties, breaking naive interpolation from smaller ladders trained under the same $K$ constraint.

Although eval losses do not scale predictably with FLOP budgets under unique data constraints, \emph{fractional loss increase} relative to the approximately-infinite baseline remains predictable:
\begin{align}
\label{eq:frac-inc}
\mathrm{FracInc}(K)
&\coloneqq
\frac{L(K)-L(\infty)}{L(\infty)}.
\end{align}
This poses a challenge for those who predict scaling behavior, since we lack an infinite-unique-data baseline.  \Cref{subsec:theory_latents} develops theory to resolve this problem and restore predictable scaling in the presence of semantic duplicates.

\section{Theory: Scale-Dependent Effective Duplicates and Restored Scaling}
\label{sec:theory}

\subsection{Semantics as Hierarchical Latents and Semantic Duplicates}
\label{subsec:theory_latents}

We model ``same meaning, different surface form'' via latent semantics and transformations.
Let $z$ denote a \emph{semantic latent} (meaning), and let $\tau$ denote a \emph{surface transformation}
(language, paraphrase, formatting, casing, etc.).
A document $x$ is generated by
\begin{equation}
z \sim p(z),\qquad \tau \sim p(\tau\mid z),\qquad x = \mathcal{G}(z,\tau).
\label{eq:theory_latent_gen}
\end{equation}
Two documents $x$ and $x'$ are \emph{semantic duplicates} if they share the same $z$ but differ in $\tau$.
This abstraction covers translations: $x=\mathcal{G}(z,\tau_{\mathrm{EN}})$ and
$x'=\mathcal{G}(z,\tau_{\mathrm{ZH}})$.

To capture compositional structure, we allow $z$ itself to be hierarchical:
\begin{equation}
z^{(0)} \to z^{(1)} \to \cdots \to z^{(L)} \to x,
\label{eq:theory_hierarchy}
\end{equation}
where $z^{(0)}$ is coarse semantics (topic/world knowledge) and $z^{(L)}$ is closest to surface form.
In this view, ``duplicates'' are not a binary dataset property: two documents can share an ancestor latent
at some depth but not others. A model that only learns shallow latents may treat translations as distinct,
while a model that learns deeper invariances collapses them to the same effective representation.

The hierarchy in Eq.~\eqref{eq:theory_hierarchy} is an abstract model of compositional structure, where coarser latents $z^{(0)}$ capture broad topics/semantics while deeper latents capture increasingly fine-grained meaning
and surface realization.
This perspective is closely related to recent theoretical models of compositional data such as the
Random Hierarchy Model (RHM), which generates examples by composing features along a tree (analogous to a grammar derivation)
and predicts scale-dependent learnability of deeper levels \citep{Cagnetta_2024}.
In our setting, increasing capability corresponds to learning deeper invariances in the latent hierarchy, which enlarges the
set of surface variants that collide into the same effective semantic latent, increasing redundancy.


Using notation from \Cref{sec:emergence}, we formalize ``duplication'' in terms of training signal rather than surface form.
Let $f_\theta$ be a language model trained by next-token prediction.

\begin{definition}[Effective duplicates]
Fix $\varepsilon\in(0,1)$.
We call $x$ and $x'$ \emph{$\varepsilon$-effective duplicates at $\theta$} if
\begin{equation}
\mathrm{sim}_\theta(x,x') \ge 1-\varepsilon.
\label{eq:theory_effective_dup}
\end{equation}
\end{definition}

This definition is explicitly model-dependent: as capability/scale increases, the relation
\eqref{eq:theory_effective_dup} can merge previously distinct examples (e.g.\ translations).


To connect semantics to gradients, we use a minimal decomposition.
Let $z=z(x)$ denote the semantic latent for $x$.
We write the per-document gradient as
\begin{equation}
g(x;\theta)
=
\underbrace{\mu(\theta)}_{\text{global}}
+
\underbrace{\delta_{z}(\theta)}_{\text{semantic}}
+
\underbrace{\xi_{x}(\theta)}_{\text{surface/idiosyncratic}},
\label{eq:theory_grad_decomp}
\end{equation}
where $\E[\delta_z]=0$ and $\E[\xi_x\mid z]=0$.
Intuitively, $\delta_z$ captures the update direction shared by all surface forms of the same meaning,
while $\xi_x$ captures surface-specific variations.

A convenient summary of semantic sensitivity at scale $s$ (parameters/compute/training time) is the fraction of gradient energy
explained by the semantic component:
\begin{equation}
\rho(s)
\coloneqq
\frac{\E\|\delta_{z}(\theta(s))\|_2^2}{\E\|g(x;\theta(s))-\mu(\theta(s))\|_2^2}
\in[0,1].
\label{eq:theory_rho_def}
\end{equation}
Under mild assumptions that the surface/idiosyncratic term $\xi_x$ is approximately isotropic and independent across
different surface forms of the same latent $z$, $\rho(s)$ controls expected gradient cosine similarity.
Concretely, for semantic duplicates $x=\mathcal{G}(z,\tau)$ and $x'=\mathcal{G}(z,\tau')$ with the same $z$,
the numerator satisfies $\E\langle g(x)-\mu,\,g(x')-\mu\rangle \approx \E\|\delta_z\|_2^2$, while the denominator is
$\E\|g(x)-\mu\|_2^2 \approx \E\|\delta_z\|_2^2+\E\|\xi_x\|_2^2$, yielding the approximation
\begin{align}
\E\big[\mathrm{sim}_{\theta(s)}(x,x')\mid z\big]\ &\approx\ \rho(s),
\\ 
\E\big[\mathrm{sim}_{\theta(s)}(x,\tilde x)\big]\ &\approx\ 0 \ \text{for unrelated }\tilde x.
\label{eq:theory_rho_cos}
\end{align}
Our gradient experiments (Section~\ref{sec:emergence}) provide direct empirical evidence that $\rho(s)$ increases with both
training progress and model capability: transformations that preserve $z$ (e.g.\ translations) become increasingly aligned in gradient space.

\subsection{Replication, Redundancy, and Effective Uniqueness}
\label{subsec:theory_reuse}

Consider training on a stream constructed by sampling \emph{with replacement} from an underlying distribution over semantic
latents $z\in\mathcal{Z}$ with mixture weights $\{w_z\}$.
In our controlled experiments (Section~\ref{sec:impact}), this corresponds to uniform sampling from a pool of $K$ unique
documents (so $w_z=1/K$), but the latent view also covers non-uniform frequencies.

A key quantity is the (Simpson) latent collision probability~\cite{simpson}
\[
p_{\mathrm{lat}} \coloneqq \Prob(z=z')=\sum_{z} w_z^2,
\]
and the associated effective latent count
\begin{equation}
K_{\mathrm{eff}} \coloneqq \frac{1}{p_{\mathrm{lat}}}=\frac{1}{\sum_z w_z^2}.
\label{eq:Keff_simpson}
\end{equation}
(When $w_z\equiv 1/K$, we have $K_{\mathrm{eff}}=K$.)

Let $x_1,\dots,x_n$ be $n$ iid draws from this mixture and define the averaged centered gradient
$\bar g_n \coloneqq \frac{1}{n}\sum_{t=1}^n (g(x_t;\theta)-\mu(\theta))$.
Assume the following simplified correlation structure consistent with Eq.~\eqref{eq:theory_grad_decomp}
\begin{equation}
\begin{split}
C(x,x')
\approx
\begin{cases}
\sigma^2 & z(x)=z(x') \text{ and } x=x',\\
\rho(s)\,\sigma^2 & z(x)=z(x') \text{ and } x\neq x',\\
0 & z(x)\neq z(x'),
\end{cases}
\end{split}
\label{eq:theory_corr_model}
\end{equation}
where $C(x,x')\equiv E\langle g(x;\theta)-\mu,\,g(x';\theta)-\mu\rangle$.

\begin{proposition}[Saturation of independent training signal]
\label{prop:theory_variance_saturation}
Under \eqref{eq:theory_corr_model} and uniform sampling over $K$ classes,
\begin{align}
\E\|\bar g_n\|_2^2
&\approx
\frac{\sigma^2}{n}
\left(1 + \rho(s)\,(n-1)\,p_{\mathrm{lat}}\right)
\\
&=
\frac{\sigma^2}{n}
\left(1 + \rho(s)\,\frac{n-1}{K_{\mathrm{eff}}}\right).
\label{eq:theory_var_mean_grad}
\end{align}
Equivalently, the averaged gradient behaves like an iid average with \emph{effective} sample size
\begin{align}
n_{\mathrm{eff}}(n,K_{\mathrm{eff}};s)
&\coloneqq
\frac{n}{1+\rho(s)\,(n-1)/K_{\mathrm{eff}}}
\ \\
&\approx
\
\min\Big\{n,\ \frac{K_{\mathrm{eff}}}{\rho(s)}\Big\}.
\label{eq:theory_neff}
\end{align}
\end{proposition}

\noindent
\textbf{Interpretation.}
When $n\ll K/\rho(s)$, redundancy is negligible and signal scales like $1/n$.
When $n\gg K/\rho(s)$, semantic redundancy dominates and the number of effectively independent update directions
saturates at $K/\rho(s)$.
Because $\rho(s)$ increases with capability (Section~\ref{sec:emergence}), the same finite-$K$ stream becomes \emph{more redundant}
for larger/stronger models, i.e.\ effective uniqueness $K/\rho(s)$ shrinks with scale.

\subsection{From Effective Reuse to a Restored Scaling Law}
\label{subsec:theory_restored_scaling}

Let $C$ denote training compute. 
Let $L(C,K_{\mathrm{eff}})$ be eval loss when sampling with replacement from an effective semantic pool size $K_{\mathrm{eff}}$,
and let $L_\infty(C)$ be the baseline with effectively infinite uniqueness (negligible repeats).
Define the normalized degradation
\begin{equation}
\Delta(C,K)
\coloneqq
\frac{L(C,K)-L_\infty(C)}{L_\infty(C)}.
\label{eq:theory_Delta_def}
\end{equation}
The redundancy picture suggests that the relevant control variable is an \emph{effective reuse ratio}
\begin{equation}
r_{\mathrm{eff}}(C,K_{\mathrm{eff}})
\coloneqq
\frac{\rho(C)\,n(C)}{K_{\mathrm{eff}}},
\label{eq:theory_reuse_ratio}
\end{equation}
where $n(C)$ is the number of documents trained on at compute $C$, and $\rho(C)$ captures semantic alignment
(Section~\ref{subsec:theory_latents}).

\textbf{Assumption (Power Law Penalty in Effective Reuse).}
Over the regime where scaling laws are measured, we posit
\begin{equation}
\Delta(C,K)\ \approx\ \lambda \, r_{\mathrm{eff}}(C,K)^{\eta}.
\label{eq:theory_power_reuse}
\end{equation}
This is a parsimonious way to encode that (i) no penalty occurs when reuse is negligible and
(ii) penalty grows smoothly with semantic redundancy.

\textbf{Compute Dependence and the Plane Law.}
Over a limited compute range, we approximate both $n(C)$ and $\rho(C)$ by power laws
\begin{equation}
n(C)\propto C^{u},
\qquad
\rho(C)\propto C^{v}.
\label{eq:theory_uv}
\end{equation}
Then \eqref{eq:theory_power_reuse} yields
\begin{equation}
\Delta(C,K_{\mathrm{eff}}) \approx\ a C^{\beta}K_{\mathrm{eff}}^{-\gamma},
\quad
\beta=\eta(u+v),\quad \gamma=\eta,
\label{eq:theory_plane_law}
\end{equation}
where $a>0$ absorbs constants.
Equation~\eqref{eq:theory_plane_law} is the \emph{minimal global scaling correction} compatible with:
(i) reuse increasing with compute ($u>0$), and (ii) semantic sensitivity increasing with compute ($v\ge 0$).
A special \emph{ratio-only} law $\Delta\propto(\sqrt{C}/K)^\eta$ corresponds to $u=1/2$ and $v=0$,
which can be too restrictive when $\rho(C)$ grows with scale.

\textbf{Restored Predictivity.}
Combining \eqref{eq:theory_Delta_def} and \eqref{eq:theory_plane_law} gives the restored loss prediction:
\begin{equation}
L_{\mathrm{pred}}(C,K_{\mathrm{eff}})
=
L_\infty(C)\Big(1+a\,C^{\beta}K_{\mathrm{eff}}^{-\gamma}\Big).
\label{eq:theory_Lpred}
\end{equation}
In our experiments (Section~\ref{sec:impact}), $L_\infty(C)$ is measured directly from the ``approximately infinite unique data''
runs at the same compute, so restoring predictivity requires fitting only $(a,\beta,\gamma)$. In~\cref{app:hutter-duplicates}, we provide a collision-aware scaling correction can be
derived by combining a Hutter-style learning curve~\cite{hutter2021learning} with an effective-sample-size reduction induced by duplicate/semantic-collision gradients.

\textbf{Empirical Validation and Minimality.}
On our controlled scaling ladders, the 3-parameter plane law \eqref{eq:theory_plane_law} accurately predicts
\emph{all} eval losses across $(C,K)$, including the breakdown regime, with small average relative error,
whereas the 2-parameter ratio-only constraint can substantially underpredict the catastrophic $K=10^5$ main run.
This supports the interpretation that semantic sensitivity $\rho(C)$ contributes nontrivially to the compute exponent $\beta$.

\subsection{Estimating an Effective Semantic Pool Size from Mean Nearest-Neighbor Cosine}
\label{subsec:keff_from_mean_nn}

In real pretraining, the ``number of unique semantic items'' $K$ is not directly observable.
However, our restored scaling law only requires an \emph{effective uniqueness}---the rate at which training samples
collide under the model's semantic resolution.
Here we show how to estimate an effective $K_{\mathrm{eff}}$ using only a \emph{mean nearest-neighbor cosine} statistic
computed from embeddings of the \emph{sampled training stream} (which includes repeats).

\textbf{Setup: Cosine is Measured on a Fixed Embedding Subsample of the Stream.}
For each training run we take a subsample of $N_{\mathrm{meas}}$ training documents from the run's data stream (including repeats),
embed each document with a fixed embedding model, and unit-normalize to obtain vectors $v_t\in\mathbb{S}^{d-1}$.
We then compute the nearest-neighbor cosine for each embedded sample
\begin{align}
M_t \coloneqq \max_{s\neq t}\langle v_t, v_s\rangle,
\quad
\overline{M}_{N_{\mathrm{meas}}} \coloneqq \frac{1}{N_{\mathrm{meas}}}\sum_{t=1}^{N_{\mathrm{meas}}} M_t .
\end{align}
All quantities below refer to this fixed measurement size $N_{\mathrm{meas}}$.
(In our controlled ladder, $N_{\mathrm{meas}}$ is constant across runs; if $N_{\mathrm{meas}}$ is not logged, it can be inferred
from the small-$K$ regime where exact repeats are frequent.)
Crucially, $\overline{M}_N$ is computed on the \emph{stream} of size $N$, which depends on $C$ (through the number of
examples processed), not on the unknown pool size $K$.

\textbf{Step 1: Background NN Similarity without Collisions.}
Let $m_0(N)$ denote the expected mean NN cosine when the nearest neighbor is \emph{not} a semantic collision
(i.e., no same-latent partner appears among the $N-1$ other samples).
In practice we estimate $m_0(N)$ from a high-uniqueness reference stream (largest-$K$ pool or without-replacement stream),
where exact repeats are negligible, using the same embedding pipeline.
\begin{figure*}[t]
    \centering
    \includegraphics[width=0.95\linewidth]{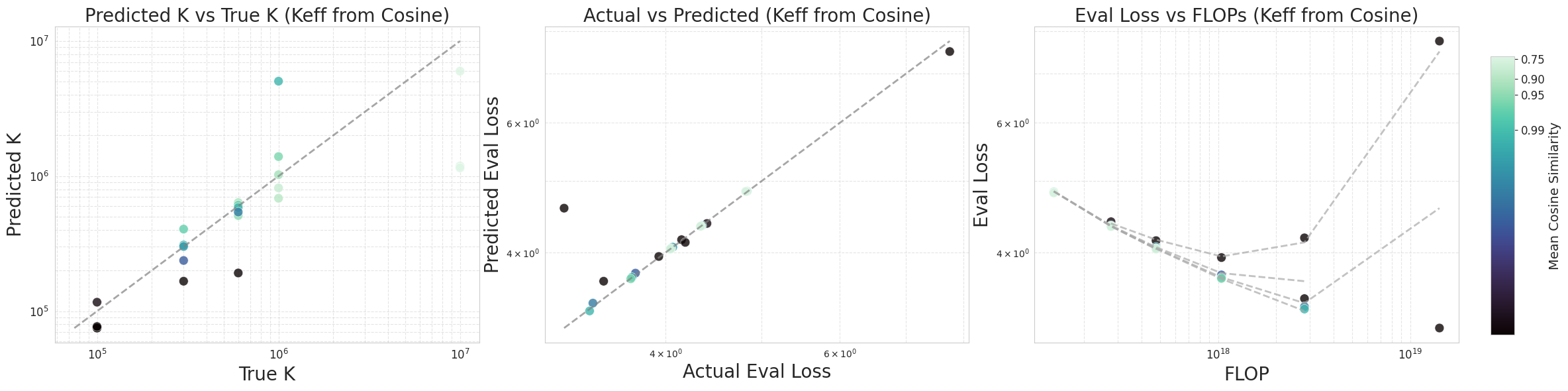}
    \caption{\textbf{Predictable scaling can be restored by accounting for limited semantic diversity:}  We use dataset size and mean cosine similarity to estimate $K$ via \Cref{eq:Keff_hat_stream} (left).  We then plug our estimate of $\widehat{K}_{\textrm{eff}}$ into  \Cref{eq:theory_Lpred} to estimate the loss (Center).  This produces scaling curves that align closely with the empirical eval losses (right).}
    \label{fig:corrected_scaling_with_k_eff_from_cos}
\end{figure*}

\textbf{Step 2: A Two-Component Model for $\overline{M}_N$.}
We model each $M_t$ as either:
(i) a background neighbor with mean $m_0(N)$, or
(ii) a collision neighbor (same latent) with typical similarity $m_+\in(m_0(N),1]$.
Let $q_N$ be the probability that a given sample has at least one collision neighbor among the other $N-1$ samples.
Then
\begin{equation}
\E[\overline{M}_N]
\approx
(1-q_N)\,m_0(N) + q_N\,m_+.
\label{eq:q_from_meanM_stream}
\end{equation}
Solving gives the estimator
\begin{equation}
\widehat q_N
\coloneqq
\mathrm{clip}\!\left(
\frac{\overline{M}_N - m_0(N)}{m_+ - m_0(N)},
\,0,\,1
\right).
\label{eq:qhat_stream}
\end{equation}
In our controlled experiment where collisions correspond to \emph{exact repeats}, we take $m_+=1$ (up to numerical precision).
For semantic (non-exact) collisions, $m_+<1$ can be calibrated using known semantic-duplicate pairs (e.g.\ translations).

\textbf{Step 3: Invert $\widehat q_N$ into an Effective Latent Count $K_{\mathrm{eff}}$.}
Let $z$ be the semantic latent with mixture weights $\{w_z\}$ and collision probability
$p_{\mathrm{lat}}=\Prob(z=z')=\sum_z w_z^2$.
Define the effective number of latents (Simpson effective size)
\begin{equation}
K_{\mathrm{eff}} \coloneqq \frac{1}{p_{\mathrm{lat}}}=\frac{1}{\sum_z w_z^2}.
\label{eq:Keff_simpson_stream}
\end{equation}
For a latent mixture with weights $\{w_z\}$, the probability that a given draw has at least one same-latent partner among the
other $N_{\mathrm{meas}}-1$ draws is
\begin{equation}
\begin{split}
q_{N_{\mathrm{meas}}}
&=
1-\sum_z w_z\,(1-w_z)^{N_{\mathrm{meas}}-1} 
\\ 
&\approx
1-\exp\!\left(-\frac{N_{\mathrm{meas}}-1}{K_{\mathrm{eff}}}\right),
\label{eq:qN_occ_stream}
\end{split}
\end{equation}
where the approximation holds when the mixture has no heavy modes (all $w_z\ll 1$); in the uniform-$K$ case it is exact up to
the standard $\log(1-x)\approx -x$ approximation, see \cref{app:derivation_qN}.
Inverting yields
\begin{equation}
\widehat K_{\mathrm{eff}}
\coloneqq
\frac{N_{\mathrm{meas}}-1}{-\log(1-\widehat q_{N_{\mathrm{meas}}})}.
\label{eq:Keff_hat_stream}
\end{equation}
\textbf{Step 4: A $K$-free Restored Scaling Law.}
Our restored degradation model is
\[
\Delta(C,K)\coloneqq\frac{L(C,K)-L_\infty(C)}{L_\infty(C)}
\approx a\,C^{\beta}K^{-\gamma}.
\]
Replacing $K$ by $\widehat K_{\mathrm{eff}}$ gives a correction depending only on observable stream geometry $\Delta(C) \approx a C^{\beta}\,\widehat K_{\mathrm{eff}}^{-\gamma}$ as 
\begin{equation}
L_{\mathrm{pred}}(C)=L_\infty(C)\bigl(1+\Delta(C)\bigr).
\label{eq:Kfree_restored_stream}
\end{equation}

\textbf{Validation on the Controlled Ladder.}
On the common evaluation set of runs in our controlled $K$-pool experiment, the plane law using the true pool size $K$
achieves mean absolute relative error $\approx 0.77\%$ (median $\approx 0.28\%$).
Replacing $K$ with $\widehat K_{\mathrm{eff}}$ estimated from mean NN cosine via
Eqs.~\Cref{eq:q_from_meanM_stream}--\eqref{eq:Keff_hat_stream} achieves $\approx 0.90\%$ (median $\approx 0.24\%$).
Thus, even with access only to a mean cosine statistic, $\widehat K_{\mathrm{eff}}$ recovers most of the predictivity of the
true-$K$ scaling correction.  See \Cref{fig:corrected_scaling_with_k_eff_from_cos}.

\begin{remark}[Identifiability from mean NN cosine]
The mapping $\overline{M}_{N_{\mathrm{meas}}}\mapsto \widehat q_{N_{\mathrm{meas}}}$ requires specifying both a background term
$m_0(N_{\mathrm{meas}})$ and a collision-similarity level $m_+$.
With only the mean NN cosine available, $m_+$ cannot be identified without external calibration; in the controlled with-replacement experiment, exact repeats imply $m_+\approx 1$, while for semantic (non-exact) collisions one can calibrate $m_+$ using known semantic-duplicate pairs (e.g.\ translations) embedded by the same model.
\end{remark}

\section{Discussion and Future Directions}

We discover an insidious source of scale-dependence that can impact the training of large language models, but not smaller language models: as model capabilities increase, training signals from semantically equivalent documents align. Thus, semantically equivalent documents in the corpora may act similarly to exact duplicates, harming model quality. Moreover, as training data scale, the number of semantic collisions increases far more quickly than one would expect based on trends gleaned from small corpora.  We model this effect on small language models and propose scaling laws that account for semantic diversity in the dataset, restoring predictable scaling.  

Our experiments have profound implications for the future of language models. Until now, industry convention has been to bet trillions of dollars on the success of the bitter lesson: scale, scale, scale, and super-intelligence will follow \citep{sutton2019bitter}. The only obstruction on this path has been the limited number of training data in web-scale corpora. Frontier labs have tried to sidestep this obstacle by synthesizing massive corpora comprised of LLM-generated text.  Our findings tell a cautionary tale about this approach: even if one can scale the raw number of tokens to asymptotically high regimes, semantic diversity may be just as important as data volume.  As we show in \Cref{fig:synthetic_diversity}, synthetic data scales poorly with respect to semantic diversity.  Our experiments emphasize the importance of seeding semantic diversity in synthetic data.  There is only one other path: if the sum total of extant semantically distinct human thoughts is insufficient to train modern LMs, then labs must invest in more data-efficient training and architectures.  We discuss limitations and future work in Appendix \ref{app:limitations_and_future_work}.


\section*{Impact Statement}
This paper presents work whose goal is to advance the field of machine learning. There are many potential societal consequences of our work, none of which we feel must be specifically highlighted here.

\section*{Acknowledgements}

SK acknowledges support from NSF 2046795 and 2205329, IES R305C240046, ARPA-H, the MacArthur Foundation, Schmidt Sciences, HAI, OpenAI, Microsoft, and Google.

We express gratitude for helpful discussions with William Meng and first-rate copy editing from Stephanie Schneider.

\clearpage

\bibliography{example_paper}
\bibliographystyle{icml2026}

\clearpage
\appendix
\onecolumn

\section{Related Work} \label{related_work}

Predictable scaling has been central to deep learning since early work on learning curves and performance prediction \citep{dohmanlearningcurves2015, hestness2017deeplearningscalingpredictable}.
Despite the analytic intractability of modern neural networks, empirical scaling laws often predict loss as a function of model size, data size, and compute with high accuracy \citep{schoenholz2017deepinformationpropagation, rosenfeld2019constructivepredictiongeneralizationerror, kaplan2020scalinglawsneurallanguage, hoffmann2022trainingcomputeoptimallargelanguage}.
Scaling predictability governs many aspects of training recipes, including parameterization, learning rates, depth-to-width ratios, initialization, warmup, and batch size
\citep{kadra2023scalinglawshyperparameteroptimization, yang2022tensorprogramsvtuning, xiong2020layernormalizationtransformerarchitecture, wang2022deepnetscalingtransformers1000, zhang2019fixupinitializationresiduallearning, kalra2024warmuplearningrateunderlying, mccandlish2018empiricalmodellargebatchtraining}.
A persistent challenge is identifying \emph{scale-dependent} factors that undermine predictable extrapolation \citep{ivgi2022scalinglawsmicroscopepredicting, schaeffer2023emergentabilitieslargelanguage, porian2025resolvingdiscrepanciescomputeoptimalscaling}. Our work highlights a new source of scale dependence linked to semantic duplicates and their frequency at web scale.

We build on work studying repeated or low-uniqueness training data. \citet{hernandez2022scalinglawsinterpretabilitylearning} showed that repeating a small subset of training examples can substantially reduce the effective parameter size predicted by scaling, and subsequent work reports that the effects of repetition can grow with scale. Because near-duplicates are common in web corpora \citep{web_scale_duplicates}, practical pipelines deploy hashing and approximate matching to identify and eliminate such ``fuzzy duplicates" \citep{simhash1997, manku_minhash, khan2025lshbloommemoryefficientextremescaledocument}.
Unlike settings with \emph{explicit} repeats (e.g., injected duplicates or many epochs) \citep{hernandez2022scalinglawsinterpretabilitylearning, muennighoff2025scalingdataconstrainedlanguagemodels, yan2025largerdatasetsrepeatedmore},
we emphasize an \emph{implicit} and \emph{scale-dependent} notion of repetition: as models become semantically sensitive, semantically equivalent documents may function as duplicates, and the prevalence of semantic collisions grows with corpus scale.

Our gradient-based measurements relate to work that treats gradients as training signals and influence proxies \citep{pruthi2020estimatingtrainingdatainfluence, koh2020understandingblackboxpredictionsinfluence}.
Our observation that semantic structure becomes more salient over training aligns with prior work on the emergence and probing of semantic representations during pretraining
\citep{jin_semantic_emergence_2024, chen2024quantifyingsemanticemergencelanguage, aljaafari2025tracetrackingemergencesemantic, wang2024probingemergencecrosslingualalignment}.
Unlike prior work that treats semantic emergence as a purely beneficial phenomenon, we connect it to a potential failure mode: semantic duplicates can create redundant training signals that disproportionately affect capable LMs.

A complementary line of work studies how neural networks learn hierarchical and compositional structure.
Recent theory introduces stylized latent-data models such as the Random Hierarchy Model (RHM), in which examples are generated by
composing features along a tree (analogous to a grammar derivation), yielding sharp predictions about which levels of the hierarchy
are learnable at a given scale \citep{Cagnetta_2024,cagnetta2025learningcurvestheoryhierarchically,cagnetta_language_learned,Sclocchi_2025}.
In language modeling, formal-language and grammar-based probes have been used to analyze whether attention-based architectures can
represent and generalize hierarchical dependencies, including theoretical limitations of self-attention \citep{hahn2020theoretical}
and empirical studies of Transformer recognition of formal languages \citep{bhattamishra2020ability}.
Most recently, \citet{schulz2025unraveling} directly characterizes how language models learn context-free grammars over training.
Our work connects to these perspectives by highlighting a distinct consequence of learning deeper invariances:
as models become semantically/compositionally sensitive, semantically equivalent documents increasingly behave as effective duplicates,
amplifying the impact of semantic collisions at corpus scale.

\begin{figure}[t]
    \centering
    \includegraphics[width=\linewidth]{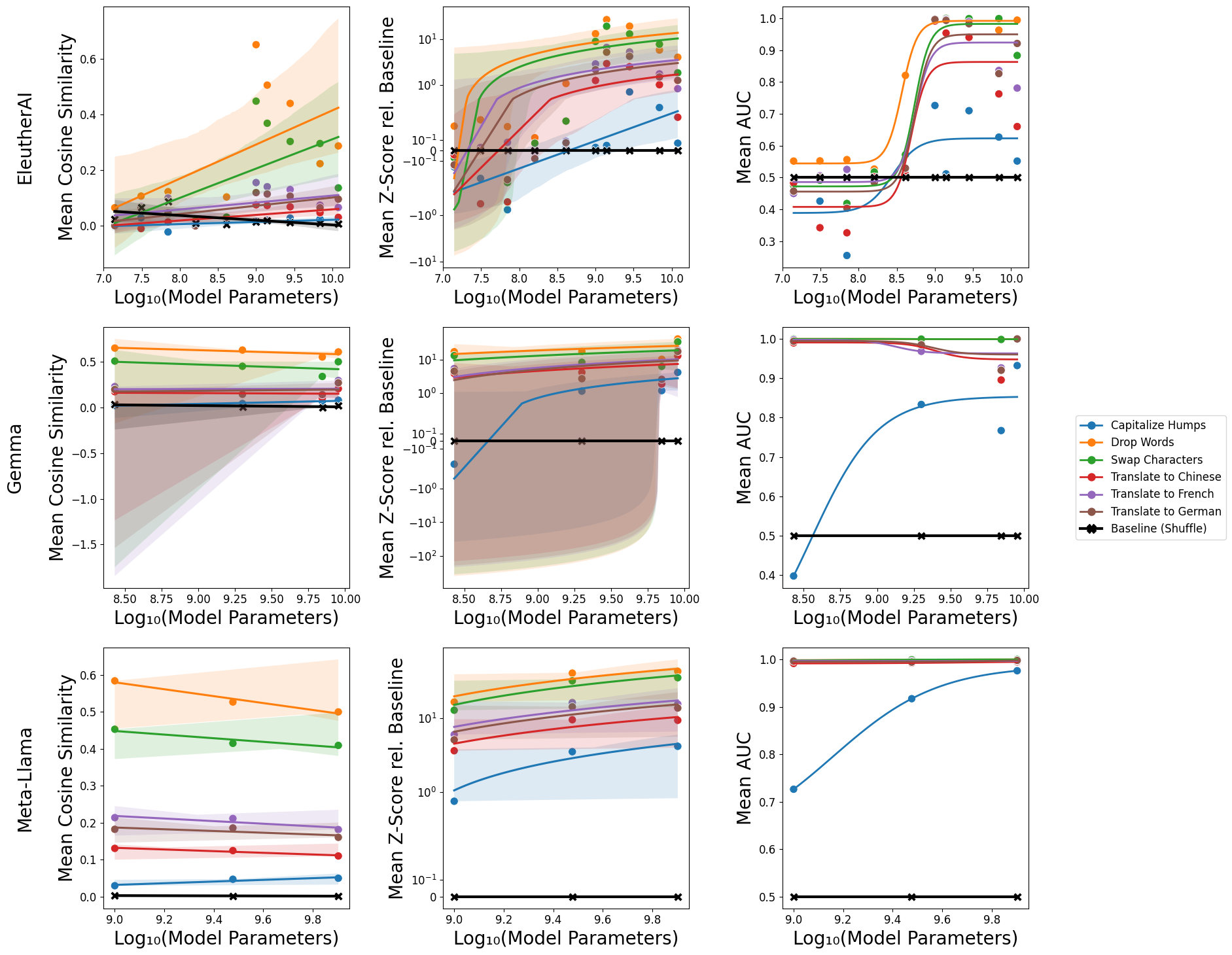}
    \caption{\textbf{Semantic-preserving transformations yield more aligned gradients for larger/stronger models:} We display the same data as in \Cref{fig:grad-sim-matrix}.}
    \label{fig:grad-sim-ind-family}
\end{figure}

\section{Limitations and Future Work} \label{app:limitations_and_future_work}

This work has several limitations.

First, because we are unable to train multi-billion-parameter models on our compute budget, we simulated the impact of semantic duplicates using exact duplicates.  We drew appropriate comparisons by basing estimates on mean cosine similarity of semantic embeddings, but behavior may differ for large models trained on semantic rather than exact duplicates.  Future work could validate our findings on large models by training on semantically deduplicated datasets. The formation of such datasets requires future research attention and resources.  

As another limitation, the semantic embeddings that we used were from EmbeddingGemma-300m. Although this is a state-of-the-art embedding model used by many frontier labs today for data exploration, it still does not produce perfectly isotropic or representative embeddings.  Training better embedding models that utilize the latent space more efficiently could improve confidence in results.

\section{Algorithm for Gradient Comparison}
\label{app:alg-grad}

\begin{algorithm}[H]
\caption{Experiment 1: Gradient similarity for semantic duplicates with negative baseline distribution}
\label{alg:exp1}
\begin{algorithmic}[1]
\Require Base texts $\{x_i\}_{i=1}^N$; transformation set $\mathcal{T}$;
models $\{f^{(k)}\}_{k=1}^K$ with checkpoints $\{\theta_{k,s}\}$;
loss $\ell$; token budget $T$; number of negative pairing rounds $R$
\Ensure For each $(k,s,\tau)$: positives $\{s_i^{+}\}$, baseline negatives $\mathcal{S}^{-}$, AUC, Z-score summary

\State Truncate each $x_i$ to at most $T$ tokens (model tokenizer)

\For{$k \gets 1$ \textbf{to} $K$}
  \ForAll{checkpoints $s$ of model $k$}
    \State Load parameters $\theta \gets \theta_{k,s}$

    \Comment{Compute gradients for all base texts once}
    \For{$i \gets 1$ \textbf{to} $N$}
      \State $g_i \gets \nabla_{\theta}\,\ell(x_i;\theta)$
    \EndFor

    \Comment{Build a negative baseline distribution from many random pairings}
    \State $\mathcal{S}^{-} \gets [\,]$
    \For{$r \gets 1$ \textbf{to} $R$}
      \State Sample pairing map $j_r(\cdot)$ such that $j_r(i)\neq i$ for all $i$
      \For{$i \gets 1$ \textbf{to} $N$}
        \State $\mathcal{S}^{-}.\mathrm{append}\!\left(\cos(g_i, g_{j_r(i)})\right)$
      \EndFor
    \EndFor
    \State Compute $\mu^{-}$ and $\sigma^{-}$ from $\mathcal{S}^{-}$

    \Comment{Evaluate each transformation on all texts}
    \ForAll{$\tau \in \mathcal{T}$}
      \For{$i \gets 1$ \textbf{to} $N$}
        \State $x_i^{(\tau)} \gets \tau(x_i)$
        \State $g_i^{(\tau)} \gets \nabla_{\theta}\,\ell(x_i^{(\tau)};\theta)$
        \State $s_i^{+} \gets \cos(g_i, g_i^{(\tau)})$
        \State $z_i \gets (s_i^{+}-\mu^{-})/\sigma^{-}$
      \EndFor
      \State Compute AUC using positives $\{s_i^{+}\}_{i=1}^N$ vs negatives $\mathcal{S}^{-}$
      \State Summarize Z-scores (e.g., mean/median over $i$)
    \EndFor
  \EndFor
\EndFor
\end{algorithmic}
\end{algorithm}

\section{Deriving The Partner Probability and the Effective Latent Count Approximation}
\label{app:derivation_qN}

This appendix derives Eq.~\eqref{eq:qN_occ_stream} step by step, starting from an exact expression for a general latent
mixture and then giving a controlled approximation that yields the compact form
$1-\exp(-(N_{\mathrm{meas}}-1)/K_{\mathrm{eff}})$.

\paragraph{Setup (Latent-Mixture Model).}
Let $Z$ be a discrete semantic latent taking values in an index set $\mathcal{Z}$ with mixture weights
$\{w_z\}_{z\in\mathcal{Z}}$, i.e.\ $\Prob(Z=z)=w_z$ and $\sum_{z} w_z = 1$.
Let $Z_1,\dots,Z_N \stackrel{iid}{\sim} \{w_z\}$ denote the latents of $N$ independent draws
(e.g.\ $N=N_{\mathrm{meas}}$ samples from the training stream).

For a given draw $i$, define the event that it has at least one same-latent partner among the other $N-1$ draws:
\[
A_i \coloneqq \{\exists j\neq i:\ Z_j = Z_i\}.
\]
By exchangeability, $\Prob(A_i)$ does not depend on $i$, so we analyze $A_1$.

\subsection{Exact Expression for $q_N$}
Define
\[
q_N \coloneqq \Prob(A_1) = \Prob\big(\exists j\neq 1:\ Z_j = Z_1\big).
\]
Condition on $Z_1=z$. Then each of the remaining $N-1$ draws matches $z$ with probability $w_z$, independently, so
the number of matches among draws $2,\dots,N$ is $\mathrm{Binomial}(N-1,w_z)$. Hence,
\begin{equation}
\Prob(\text{no partner for draw 1}\mid Z_1=z)
= \Prob(Z_2\neq z,\dots,Z_N\neq z\mid Z_1=z)
= (1-w_z)^{N-1}.
\end{equation}
Averaging over $Z_1$ gives the exact identity
\begin{equation}
\Prob(\text{no partner for draw 1})
=
\sum_{z\in\mathcal{Z}} \Prob(Z_1=z)\,(1-w_z)^{N-1}
=
\sum_{z} w_z(1-w_z)^{N-1}.
\end{equation}
Therefore,
\begin{equation}
\boxed{
q_N
=
1-\sum_{z} w_z(1-w_z)^{N-1}.
}
\label{eq:qN_exact}
\end{equation}
This is the first line of Eq.~\eqref{eq:qN_occ_stream} and is \emph{exact} for any discrete mixture.

\subsection{From Mixture Weights to $K_{\mathrm{eff}}$}
A key quantity is the probability that \emph{two independent draws} share the same latent:
\begin{equation}
p_{\mathrm{lat}}
\coloneqq
\Prob(Z=Z')
=
\sum_{z} \Prob(Z=z)\Prob(Z'=z)
=
\sum_{z} w_z^2.
\label{eq:plat_def}
\end{equation}
This is the Simpson collision probability.
It induces the \emph{Simpson effective number of latents}
\begin{equation}
\boxed{
K_{\mathrm{eff}} \coloneqq \frac{1}{p_{\mathrm{lat}}} = \frac{1}{\sum_z w_z^2}.
}
\label{eq:Keff_def}
\end{equation}
In the uniform-$K$ case ($w_z=1/K$ for $z=1,\dots,K$), we have $p_{\mathrm{lat}}=1/K$ and thus $K_{\mathrm{eff}}=K$.

\subsection{Approximation: Rare-Collision / No-Heavy-Modes Regime}
We now explain the approximation
\[
q_N \approx 1-\exp\!\left(-(N-1)\sum_z w_z^2\right)
= 1-\exp\!\left(-\frac{N-1}{K_{\mathrm{eff}}}\right).
\]

\paragraph{Step 1: Poissonizing the Binomial for Small $w_z$.}
For small $w_z$, the binomial $\mathrm{Binomial}(N-1,w_z)$ is well-approximated by
$\mathrm{Poisson}(\lambda_z)$ with rate $\lambda_z=(N-1)w_z$.
In particular,
\begin{equation}
(1-w_z)^{N-1}
=
\exp\!\big((N-1)\log(1-w_z)\big)
=
\exp\!\left(-(N-1)w_z + O\big((N-1)w_z^2\big)\right),
\label{eq:binom_to_exp}
\end{equation}
so when $\max_z w_z \ll 1$ and $(N-1)\max_z w_z^2$ is not too large, we may use
\begin{equation}
(1-w_z)^{N-1} \approx \exp\!\big(-(N-1)w_z\big).
\label{eq:binom_poisson_approx}
\end{equation}
Plugging \eqref{eq:binom_poisson_approx} into \eqref{eq:qN_exact} yields
\begin{equation}
q_N
\approx
1-\sum_z w_z \exp\!\big(-(N-1)w_z\big).
\label{eq:qN_mixture_exp}
\end{equation}

\paragraph{Step 2: Collapsing the Mixture to a Single Effective rate.}
Let $W$ be the random variable $W\coloneqq w_{Z_1}$ when $Z_1\sim\{w_z\}$, i.e.\ $\Prob(W=w_z)=w_z$.
Then \eqref{eq:qN_mixture_exp} can be written compactly as
\begin{equation}
\sum_z w_z \exp\!\big(-(N-1)w_z\big)
= \E\!\left[e^{-(N-1)W}\right].
\label{eq:mgfW}
\end{equation}
Moreover,
\[
\E[W] = \sum_z w_z \cdot w_z = \sum_z w_z^2 = p_{\mathrm{lat}} = \frac{1}{K_{\mathrm{eff}}}.
\]
If the mixture has \emph{no heavy modes} (informally: $w_z\ll 1$ and the distribution of $W$ is not extremely spread out),
we can approximate the expectation in \eqref{eq:mgfW} by its mean-field form:
\begin{equation}
\E\!\left[e^{-(N-1)W}\right]
\approx
\exp\!\big(-(N-1)\E[W]\big)
=
\exp\!\left(-(N-1)\sum_z w_z^2\right)
=
\exp\!\left(-\frac{N-1}{K_{\mathrm{eff}}}\right).
\label{eq:meanfield_exp}
\end{equation}
A standard way to justify \eqref{eq:meanfield_exp} is via a cumulant (Taylor) expansion:
\[
\log \E[e^{-aW}]
=
-a\,\E[W] + \frac{a^2}{2}\mathrm{Var}(W) + O\!\big(a^3\E[|W-\E W|^3]\big),
\qquad a \coloneqq N-1,
\]
so if $a^2\mathrm{Var}(W)$ is small compared to $a\E[W]$ (i.e.\ $W$ is concentrated around its mean at the scale
relevant for $a$), then $\log \E[e^{-aW}]\approx -a\E[W]$ and \eqref{eq:meanfield_exp} follows.

\paragraph{Putting the Steps Together.}
Combining \eqref{eq:qN_mixture_exp} with \eqref{eq:meanfield_exp} yields
\begin{equation}
\boxed{
q_N
\approx
1-\exp\!\left(-(N-1)\sum_z w_z^2\right)
=
1-\exp\!\left(-\frac{N-1}{K_{\mathrm{eff}}}\right).
}
\label{eq:qN_occ_stream_app}
\end{equation}
This matches Eq.~\eqref{eq:qN_occ_stream} in the main text.

\subsection{Sanity Check: uniform-$K$ case}
If $w_z=1/K$ for $z=1,\dots,K$, then \eqref{eq:qN_exact} becomes
\[
q_N
=
1-\sum_{z=1}^K \frac{1}{K}\left(1-\frac{1}{K}\right)^{N-1}
=
1-\left(1-\frac{1}{K}\right)^{N-1},
\]
and using $\log(1-x)\approx -x$ gives
\[
q_N \approx 1-\exp\!\left(-\frac{N-1}{K}\right).
\]
Since $K_{\mathrm{eff}}=K$ in the uniform case, Eq.~\eqref{eq:qN_occ_stream_app} recovers the standard occupancy approximation
exactly up to the usual $\log(1-x)\approx -x$ step.

\subsection{Remark: What Breaks when there are Heavy Modes?}
If some $w_z$ are not small (a few ``heavy'' semantics), then:
(i) the Poisson approximation \eqref{eq:binom_poisson_approx} can be inaccurate for those modes, and
(ii) the mean-field collapse \eqref{eq:meanfield_exp} can be poor because $W$ is no longer concentrated.
In that case, Eq.~\eqref{eq:qN_exact} remains correct and can be used directly, and $K_{\mathrm{eff}}$
still meaningfully summarizes pairwise collision probability via \eqref{eq:Keff_def}, but the single-exponential
approximation to $q_N$ may systematically overestimate collision probability.

\section{A First-Principles Model of Duplicate-Limited Scaling via Hutter-Style Learning Curves}
\label{app:hutter-duplicates}

This appendix derives a collision-aware scaling correction by combining:
(i) a Hutter-style learning-curve model in which performance improves as a power law in the number of
\emph{independent} training signals, and
(ii) a reduction of independent signal due to duplicates/semantic collisions.
The goal is not a fully realistic theory of language modeling, but a minimal mechanism that explains why a
plane law of the form $\Delta(C,K)\approx a\,C^\beta K^{-\gamma}$ arises naturally.

\paragraph{Step 1: A Hutter-Style ``New Information'' Learning Curve.}
A classic abstraction (learning curve theory) models learning progress as driven by discovering previously unseen
``features'' or ``types'' in a heavy-tailed environment.
Concretely, let $z$ denote a latent ``type'' (semantic class, rule, or pattern) with weights $\{w_z\}$.
Consider the idealized memorization learner that, upon seeing \emph{one} example of type $z$, can thereafter predict $z$
perfectly, while unseen types incur a fixed excess loss.
In this model, the expected excess risk after $n$ iid draws is proportional to the probability mass of unseen types:
\begin{equation}
\label{eq:hutter-unseen-mass}
\epsilon(n)
=
\sum_z w_z\,(1-w_z)^n,
\end{equation}
a form that appears in learning-curve theory and is closely related to occupancy/species discovery.
(For a detailed treatment and conditions under which heavy tails yield power laws, see \citet{hutter2021learning}.) 

\paragraph{Step 2: Power Laws from Heavy Tails.}
If the type weights follow a Zipf/regularly varying tail, $\epsilon(n)$ follows a power law:
\begin{equation}
\epsilon(n)\;\propto\; n^{-\alpha}
\quad\text{for some }\alpha\in(0,1),
\label{eq:hutter-powerlaw}
\end{equation}
with $\alpha$ determined by the tail index of $\{w_z\}$ (see \citet{hutter2021learning}).
We use \eqref{eq:hutter-powerlaw} as a generic ``first-principles'' justification for a power law dependence of excess loss
on the amount of \emph{independent} training signal.

\paragraph{Step 3: Duplicates Reduce the Effective Number of Independent Signals.}
In our setting, training examples are not independent sources of new information: duplicates (exact or semantic)
induce correlated gradients and therefore reduce the number of effectively independent update directions.
Let $n$ denote the number of training documents processed.
Let $K$ denote the number of effective semantic classes available (or $K_{\mathrm{eff}}$ in the main text).
Let $\rho\in[0,1]$ summarize semantic sensitivity (gradient alignment within a class) as in Eq.~\eqref{eq:theory_rho_def}.
Under the correlation model in Eq.~\eqref{eq:theory_corr_model}, Proposition~\ref{prop:theory_variance_saturation}
implies an effective sample size
\begin{equation}
\label{eq:neff-app}
n_{\mathrm{eff}}
=
\frac{n}{1+\rho\,\frac{n-1}{K}}
\;\approx\;
\frac{n}{1+r_{\mathrm{eff}}},
\qquad
r_{\mathrm{eff}}:=\rho\,\frac{n}{K}.
\end{equation}
Intuitively, $r_{\mathrm{eff}}$ is an \emph{effective reuse ratio}: when $r_{\mathrm{eff}}\ll 1$ the stream is mostly novel,
and when $r_{\mathrm{eff}}\gg 1$ the stream is dominated by redundant semantics.

\paragraph{Step 4: Substitute $n_{\mathrm{eff}}$ into the Learning Curve.}
Assume the excess loss (or excess cross-entropy) is a power law in the \emph{independent} signal count:
\begin{equation}
\label{eq:loss-vs-neff}
L(n,K)-L_\star
\;\approx\;
B\,n_{\mathrm{eff}}^{-\alpha},
\end{equation}
where $L_\star$ is an irreducible floor and $B>0$.
For the high-uniqueness baseline (negligible collisions), $n_{\mathrm{eff}}\approx n$ and
$L_\infty(n)-L_\star\approx Bn^{-\alpha}$.
For finite $K$, combining \eqref{eq:neff-app}--\eqref{eq:loss-vs-neff} gives
\begin{equation}
\label{eq:loss-ratio-app}
L(n,K)-L_\star
\;\approx\;
B\,n^{-\alpha}(1+r_{\mathrm{eff}})^{\alpha}.
\end{equation}

\paragraph{Step 5: A Duplicate-Induced Degradation Law.}
Define the normalized degradation $\Delta$ as in Eq.~\eqref{eq:theory_Delta_def}:
$\Delta:=(L(n,K)-L_\infty(n))/L_\infty(n)$.
Using \eqref{eq:loss-ratio-app} and $L_\infty(n)=L_\star+Bn^{-\alpha}$, we obtain
\begin{equation}
\label{eq:Delta-reuse-general}
\Delta(n,K)
\;\approx\;
\frac{Bn^{-\alpha}\big((1+r_{\mathrm{eff}})^\alpha-1\big)}{L_\star+Bn^{-\alpha}}.
\end{equation}
In the regime where $Bn^{-\alpha}$ is not negligible relative to $L_\star$ (typical for the losses in our controlled ladders),
the prefactor is slowly varying and \eqref{eq:Delta-reuse-general} is well-approximated by a power law in $r_{\mathrm{eff}}$.
In particular, when $r_{\mathrm{eff}}\lesssim 1$ we can linearize:
\begin{equation}
\label{eq:Delta-linear-reuse}
\Delta(n,K)
\;\approx\;
\tilde \lambda \, r_{\mathrm{eff}}
\;=\;
\tilde\lambda\,\rho\,\frac{n}{K},
\end{equation}
where $\tilde\lambda$ absorbs the slowly varying ratio in \eqref{eq:Delta-reuse-general}.
Equation \eqref{eq:Delta-linear-reuse} recovers the main-text intuition that degradation is (approximately) proportional to an
effective reuse ratio.

\paragraph{Step 6: Translating to Compute and the Plane Law.}
Let $C$ denote compute.
Over restricted ranges, it is empirically accurate to approximate
\[
n(C)\propto C^{u},
\qquad
\rho(C)\propto C^{v},
\]
as in Eq.~\eqref{eq:theory_uv}.
Substituting into \eqref{eq:Delta-linear-reuse} yields
\begin{equation}
\Delta(C,K)
\;\approx\;
a\,C^{u+v}\,K^{-1},
\label{eq:plane-derived-linear}
\end{equation}
which is a \emph{plane law} in $(\log C,\log K)$ with $\gamma\approx 1$.
More generally, if one does not linearize \eqref{eq:Delta-reuse-general}, the same substitution yields a plane
$\Delta(C,K)\propto C^{\beta}K^{-\gamma}$ with $\beta=\eta(u+v)$ and $\gamma=\eta$ for some effective exponent $\eta$,
matching Eq.~\eqref{eq:theory_plane_law}.

\paragraph{Discussion: Why $\rho(C)$ Should Grow with Scale (and why a Power Law is a Reasonable Local Model).}
The parameter $\rho(C)$ captures the fraction of gradient energy explained by invariances to surface form
(Eq.~\eqref{eq:theory_rho_def}).
A growing body of theory and empirical work on \emph{hierarchical/compositional} data suggests that neural networks learn
coarse, high-level structure before finer structure, and that deeper invariances require more data/compute.
For example, the random hierarchy model (RHM) formalizes language-like hierarchical generation and yields staged learning
dynamics where progressively deeper variables become learnable as sample size increases \citep{Cagnetta_2024,cagnetta2024structurelanguage}.
Separately, work on formal-language recognition by transformers highlights a connection between model depth/recurrence and
the ability to represent hierarchical (context-free) structure, which is a canonical form of compositional invariance
\citep{hahn2020computationalpower,jerad2026contextfree}.
Taken together, these results motivate modeling $\rho(C)$ as monotone increasing with scale; over the narrow compute ranges
used in scaling ladders, a power law approximation $\rho(C)\propto C^v$ is a parsimonious local model.

\paragraph{A Reduced-Parameter Variant.}
Equation \eqref{eq:plane-derived-linear} suggests a two-degree-of-freedom correction:
fix $\gamma=1$ and fit only $(a,\beta)$ (or even fit $v$ with $u$ known from the compute-to-sample mapping).
In our controlled ladders, the fitted $\gamma$ is close to $1$, consistent with this linear-reuse regime.

\newpage

\begin{figure}
    \centering
    \includegraphics[width=\linewidth]{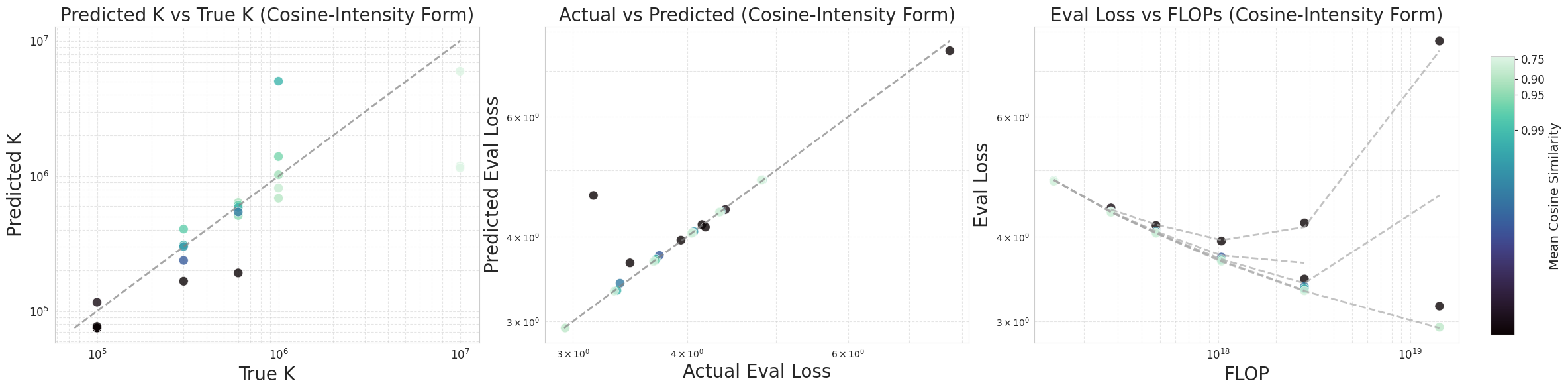}
    \caption{\textbf{Predictions of eval loss using cosine intensity-based estimation of $\widehat{K}_{\textrm{eff}}$.}}
    \label{fig:placeholder-1}
\end{figure}

\begin{figure}
    \centering
    \includegraphics[width=\linewidth]{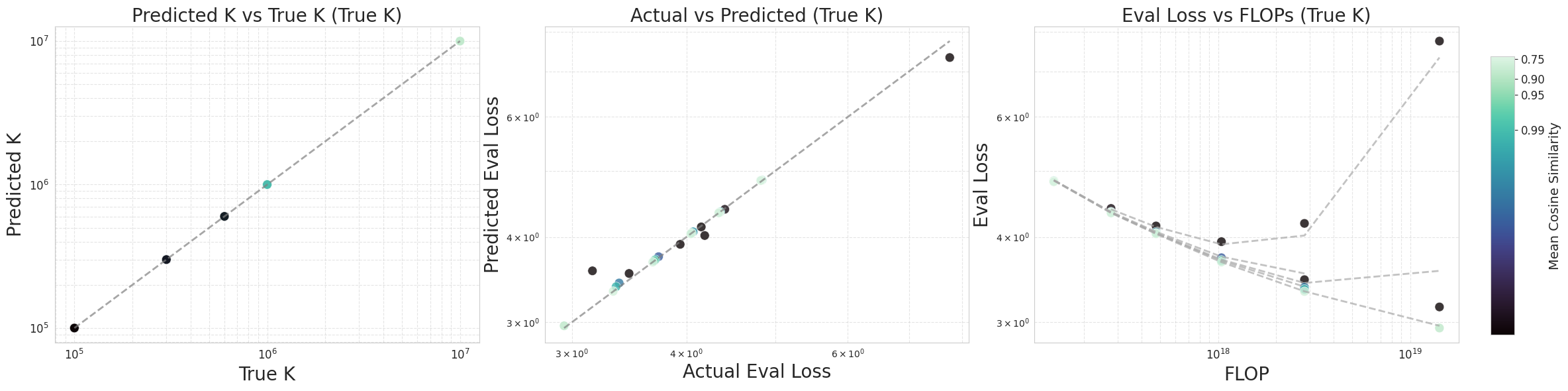}
    \caption{\textbf{Predictions of eval loss using true $K$.}}
    \label{fig:placeholder-2}
\end{figure}

\end{document}